\documentclass[letterpaper, 10 pt, conference]{ieeeconf}  

\usepackage{amsmath}   
\usepackage{amssymb}   
\usepackage{graphicx}  
\usepackage{caption}
\usepackage{booktabs}
\usepackage{multirow} 
\usepackage{subcaption}

\usepackage{algorithm} 
\usepackage{algorithmic}

\usepackage{marvosym}

\IEEEoverridecommandlockouts                              

\title{
\large \bf
A Hybrid Mamba for Audio-Visual Navigation$\dagger$
}

\author{
    Yi Wang$^{1,2,3}$, and Yinfeng Yu$^{1,2,3}$$^{,\mbox{\Letter}}$%
    \thanks{\small $\dagger$This research was financially supported by the National Natural Science Foundation of China (Grant No. 62463029).}
    \thanks{\small $^1$School of Computer Science and Technology, Xinjiang University, Urumqi 830017, China.}
    \thanks{\small $^2$Joint Research Laboratory for Embodied Intelligence, Xinjiang University.}
    \thanks{\small $^3$Joint International Research Laboratory of Silk Road Multilingual Cognitive Computing, Xinjiang University.}
    \thanks{\small $^{\mbox{\Letter}}$Yinfeng Yu is the corresponding author (Email: yuyinfeng@xju.edu.cn).}%
}

\begin{document}

\maketitle

\thispagestyle{empty}
\pagestyle{empty}

\begin{abstract}

Since the paradigm centered on convolutional neural networks and recurrent architectures was established in 2020, the fundamental backbone networks for audio-visual navigation have undergone no essential changes for more than five years, making them inadequate to support efficient representation of dynamic multimodal sequences.
This paper proposes Samba(A Hybrid Mamba for Audio-Visual Navigation). It uses the adaptive selection-enabled Mamba State Encoder (M-SE) to replace conventional GRUs for temporal aggregation, and constructs an Audio Mamba Encoder (AME) to remedy the limitations of convolutional operators in capturing global time-frequency dependencies in spectrograms.
Experiments demonstrate that Samba exhibits exceptional generalization performance when facing unheard sound sources and unseen scenes. On the Matterport3D dataset, it improves the navigation success rate (SR) by 11.3\% compared with existing state-of-the-art models, and the performance gain is even more pronounced on the Replica dataset, which features finer scene structures. Such modernized architectural reconstruction unlocks stronger embodied representation capabilities at a lower computational cost, thereby providing a highly robust technical pathway for paradigm evolution in the field of audio-visual navigation.

\end{abstract}

\section{INTRODUCTION}

Audio-visual navigation (AVN)~\cite{soundspaces,gan2020look} requires agents to localize sound sources in unmapped environments by coupling local visual geometry with long-range auditory guidance, a task heavily relying on the backbone architecture for multimodal representation and temporal state tracking. Despite recent progress, mainstream frameworks like AV-WaN~\cite{soundspaces,Waypoint} still suffer from representation mismatches due to traditional paradigms: their gated recurrent units utilize non-selective state updates that dilute critical acoustic signals with redundant environmental observations, while their convolutional neural networks introduce a local bias that disrupts the global time-frequency topological dependencies of spectrograms. These backbone-level representational limitations have become a substantial bottleneck restricting further breakthroughs in navigation performance.

To address the above limitations, and inspired by the outstanding performance of structured state-space models~\cite{Mamba} in sequence modeling, this paper constructs a hybrid state-space architecture named Samba. This framework establishes a heterogeneous hybrid modeling paradigm driven by Mamba~\cite{Mamba}, aiming to enhance the system’s representation capability by introducing a data-adaptive selective scanning mechanism. We design the Mamba State Encoder to replace traditional gated recurrent units, enabling accurate extraction and preservation of critical navigation states within complex sequences. Meanwhile, we build a bidirectional Audio Mamba Encoder, whose global modeling property resolves the time-frequency fragmentation issue in spectrogram analysis encountered by conventional convolutional operators. Furthermore, this framework strategically retains convolutional architectures in the visual branch to leverage the advantages of spatial inductive bias. Ultimately, efficient interpretation of multimodal signals is achieved through Mamba’s linear complexity and content-aware characteristics.

In summary, our main contributions are as follows:

\begin{itemize}
    \item We establish a hybrid state-space architecture named Samba, representing the first attempt to introduce state-space models as a backbone network in the field of audio-visual navigation.

    \item We design an adaptive selection-aware Mamba State Encoder and a bidirectional Audio Mamba Encoder, which address the representation dilution issue in state modeling and the structural deficiency in acoustic features, respectively.

    \item Extensive experiments validate that Samba achieves significant improvements in navigation success rate while maintaining parameter efficiency, outperforming all existing state-of-the-art baseline models.
\end{itemize}

\section{Related Work}

\begin{figure*}[ht]
    \centering
    \includegraphics[width=1\linewidth]{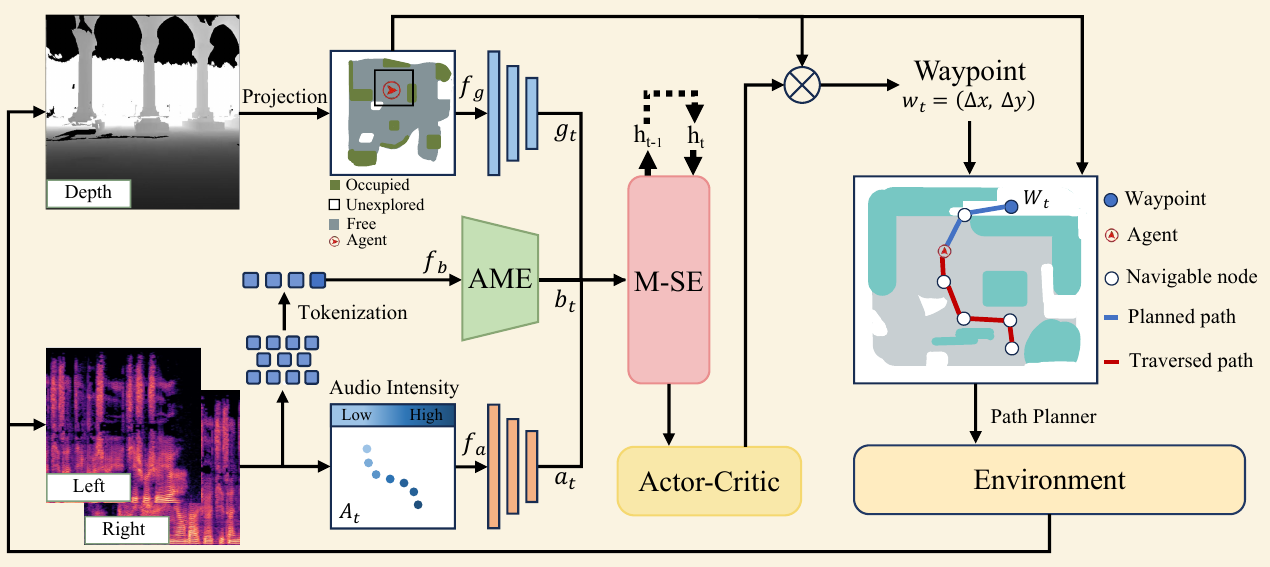}
    \caption{\textbf{Overall Architecture of the Samba Framework}. }
    \label{m1}
\end{figure*}

\subsection{Audio-Visual Navigation}
The AVN task~\cite{soundspaces,gan2020look} requires an agent to localize sound sources in complex 3D environments by jointly analyzing visual observations and binaural audio signals. Early research mainly focused on geometry-driven navigation strategies~\cite{GeoDr}, while in recent years, the end-to-end paradigm based on deep reinforcement learning has become mainstream in this field. Within this paradigm, milestone works such as AV-WaN~\cite{Waypoint,Yuttt,yu2025dope,SAAVN,CAVEN,ORAN,AI2-THOR,yu2023measuring} have established a standard backbone architecture that fuses audio, vision, and historical states. However, although these methods have achieved progress on standard simulation platforms such as SoundSpaces~\cite{soundspaces,habitat,habitat2.0}, their underlying architectures often suffer from a trade-off between representation efficiency and computational overhead. Most existing studies~\cite{CRFN,FSAVVN,li2025audio} focus on improving multimodal fusion mechanisms. In contrast, this work goes further and attempts to enhance the perception and reasoning efficiency of the navigation system by reshaping the backbone network at a fundamental level~\cite{yu2025dgfnet,yu2025dynamic,yang2026beyond,zhang2025advancing,wang2025modality,cao2024vnet,fu2025fsdenet,mattursun2024bss,zhang2024nonlinear,zhang2025iterative}.

\subsection{State Space Models and Mamba}
State Space Models (SSMs)~\cite{Mamba}, as an emerging deep learning architecture, are gradually becoming a strong alternative to Recurrent Neural Networks (RNNs)~\cite{RNN} and Transformers~\cite{attention} due to their superior performance in handling extremely long-sequence tasks. From the early Structured State Space model~\cite{Mamba} to the recent Mamba~\cite{Mamba} architecture, such models have successfully captured complex sequential dependencies while maintaining linear time complexity by introducing the Selective Scan Mechanism. In the fields of embodied intelligence and computer vision, Mamba~\cite{Mamba} has demonstrated the potential to outperform traditional architectures, enabling efficient autoregressive inference with extremely low memory overhead. The research motivation of this paper stems precisely from this: to address the long-standing state aggregation bottleneck in navigation tasks by leveraging the hardware-aware efficiency of Mamba~\cite{Mamba}.

\section{Proposed Method}

\subsection{Framework Overview}

Samba is an end-to-end system designed specifically for audio-visual navigation, which restructures perceptual encoding and state modeling operators using hierarchical selective state spaces.
To ensure seamless integration in heterogeneous environments and flexibility in policy sampling, the underlying system employs a lightweight Mamba~\cite{Mamba} operator library built upon standard PyTorch~\cite{torch} operators.
As shown in Figure~\ref{m1}, at navigation step $t$, the agent takes a depth map, binaural spectrogram, and global geometric map $G_t$ as inputs.
In the perception layer, the Audio Mamba Encoder(AME) extracts acoustic features $b_t$ with selective context through tokenization, which are then fed into the Mamba State Encoder (M-SE) together with the geometric vector $g_t$ derived from the depth map and the acoustic map feature $a_t$.
Leveraging linear recurrence, M-SE updates the historical state $h_{t-1}$ to the current global state $h_t$, enabling high-fidelity modeling of long-range environmental characteristics.
Subsequently, the Actor-Critic-based prediction network uses the PPO algorithm to sample non-myopic waypoints $w_t = (\Delta x, \Delta y)$ over the action map $W_t$.
Finally, these waypoints drive the path planner to guide the agent to accurately reach the target sound source in unknown environments.

\begin{figure*}[t]
    \centering
    \includegraphics[width=0.6\linewidth]{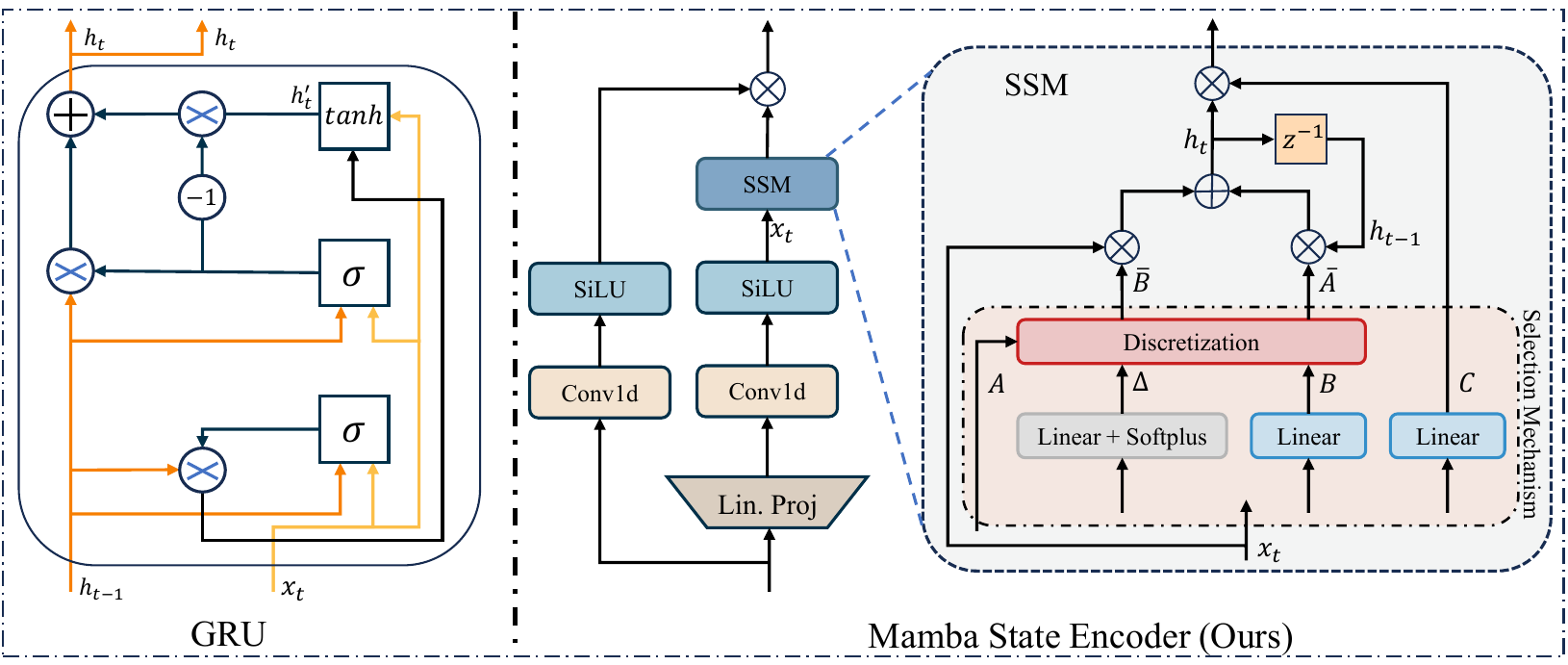}
    \caption{\textbf{Architectural Comparison between the baseline GRU and our proposed Mamba State Encoder (M-SE)}. The left side illustrates the classic Gated Recurrent Unit, while the right side details the Audio Mamba State Encoder designed in this paper.}
    \label{m3}
\end{figure*}

\subsection{Audio Mamba Encoder}

\begin{algorithm}[!t] 
\caption{Architectural Procedure of Audio Mamba Encoder}
\label{alg1} 
\small 
\begin{algorithmic}[1]
    \REQUIRE Audio spectrogram $S_t \in \mathbb{R}^{B \times C \times F \times T}$
    \ENSURE Audio state vector $b_t \in \mathbb{R}^{d_{out}}$
    
    \STATE $B, C, F, T \leftarrow \text{shape of } S_t$
    \STATE $x \leftarrow \text{Reshape}(\text{Permute}(S_t, (0, 3, 2, 1)), (B, T, C \cdot F))$ 
    \COMMENT{Tokenization and spectral flattening}
    
    \STATE $x \leftarrow \text{ReLU}(\text{LayerNorm}(\mathbf{W}_{p} x))$ 
    \COMMENT{Project features to $d_{model}$}
    
    \STATE $x \leftarrow x + \mathbf{P}[:, :T, :]$ 
    \COMMENT{Temporal positional encoding injection}
    
    \FOR{$i = 0$ \TO $N_{layers} - 1$}
        \IF{$i \pmod 2 == 1$}
            \STATE $x \leftarrow \text{Flip}(x, \text{dim}=1)$ 
            \COMMENT{Sequence reversal}
            \STATE $x \leftarrow \text{MambaLayer}_i(x)$ 
            \COMMENT{Backward selective scanning}
            \STATE $x \leftarrow \text{Flip}(x, \text{dim}=1)$ 
            \COMMENT{Restore temporal order}
        \ELSE
            \STATE $x \leftarrow \text{MambaLayer}_i(x)$ 
            \COMMENT{Forward selective scanning}
        \ENDIF
    \ENDFOR
    
    \STATE $f_b \leftarrow \text{Mean}(\text{LayerNorm}(x), \text{dim}=1)$ 
    \COMMENT{Global aggregate pooling}
    
    \STATE $b_t \leftarrow \mathbf{W}_{h} f_b$ 
    \COMMENT{Final audio state projection}
    
    \RETURN $b_t$
\end{algorithmic}
\end{algorithm}

To address the lack of global context caused by limited local receptive fields in traditional convolutional architectures when processing complex acoustic observations, we design an Audio Mamba Encoder (AME) based on selective state-space models. The core logic of this module is to elevate the audio feature extraction process from local operator transformations to global selective scanning, thereby accurately capturing the cross-time-frequency directional characteristics embedded in the binaural spectrogram $S_t \in \mathbb{R}^{C \times F \times T}$. As shown in Algorithm 1, in the feature tokenization stage, the system first linearly merges the channel and frequency dimensions of the original spectrogram via dimension permutation and rearrangement to construct a discrete token sequence. Specifically, for each time step $\tau \in \{1, \dots, T\}$, we flatten its cross-dimensional features and project them into the hidden space, yielding the initial feature vector $x_\tau$ as:

\begin{equation}
x_\tau = \operatorname{ReLU}\bigl( \operatorname{LN}( \mathbf{W}_p \cdot [S_{1,1,\tau}, \dots, S_{C,F,\tau}]^\top + \mathbf{b}_p ) \bigr),
\end{equation}
where $\mathbf{W}_p \in \mathbb{R}^{d_{model} \times (C \times F)}$ denotes the learnable linear projection matrix. To compensate for the temporal position information that may be lost when state-space models process sequences, we introduce a positional embedding matrix $\mathbf{P}$ and convert the projected sequence into a hidden input with temporal features $\mathbf{z}_\tau = x_\tau + \mathbf{P}_\tau$.

The inference logic of AME further leverages a bidirectional selective scan mechanism to enhance representations. Given that the interaural time difference and interaural level difference, on which sound source localization relies, exhibit strong temporal sensitivity, we abandon traditional static convolutional operators and adopt a heterogeneous bidirectional Mamba architecture instead. At each operator layer $i$, the model models long-range dependencies in both forward and backward directions by alternately flipping the time axis of the sequence, whose layer-wise output $\mathbf{y}_i$ can be formulated as:

\begin{equation}
\mathbf{y}_i = 
\begin{cases} 
\operatorname{MambaLayer}_i(\mathbf{z}) & \text{if } i \text{ is even} \\
\operatorname{Flip}\bigl( \operatorname{MambaLayer}_i( \operatorname{Flip}(\mathbf{z}) ) \bigr) & \text{if } i \text{ is odd}
\end{cases}
\end{equation}

This design ensures that the model can simultaneously parse the causal and anticausal correlations of audio signals in the time domain, achieving high-fidelity analysis of the sound field in complex static indoor scenes. Compared with parameter-redundant CNNs~\cite{CNNintroduction}, AME can adaptively adjust the representation weights according to the intensity distribution of the current acoustic input. Finally, the sequence features after global average pooling are mapped into the audio state vector $b_t$ by a linear head, thereby providing high-quality acoustic semantic support for the subsequent global state aggregation.

\subsection{Mamba State Encoder}

\begin{table*}[t]
\centering
\caption{Performance comparison with other methods under the Depth setting. SPL, SR, SNA are percentages.}
\label{table1}
\renewcommand{\arraystretch}{1.3} 
\setlength{\tabcolsep}{6pt}       

\begin{tabular*}{\textwidth}{@{\extracolsep{\fill}} l|ccc|ccc|ccc|ccc @{}}
\toprule
& \multicolumn{6}{c|}{\textbf{Replica}} 
& \multicolumn{6}{c}{\textbf{Matterport3D}} \\
\cmidrule(lr){2-7} \cmidrule(lr){8-13}
\textbf{Method} &
\multicolumn{3}{c|}{\textbf{Heard}} 
& \multicolumn{3}{c|}{\textbf{Unheard}} 
& \multicolumn{3}{c|}{\textbf{Heard}} 
& \multicolumn{3}{c}{\textbf{Unheard}} \\
\cmidrule(lr){2-4} \cmidrule(lr){5-7} \cmidrule(lr){8-10} \cmidrule(lr){11-13}
& \textbf{SPL}$\uparrow$ & \textbf{SR}$\uparrow$ & \textbf{SNA}$\uparrow$
& \textbf{SPL}$\uparrow$ & \textbf{SR}$\uparrow$ & \textbf{SNA}$\uparrow$
& \textbf{SPL}$\uparrow$ & \textbf{SR}$\uparrow$ & \textbf{SNA}$\uparrow$
& \textbf{SPL}$\uparrow$ & \textbf{SR}$\uparrow$ & \textbf{SNA}$\uparrow$ \\
\midrule
Random~\cite{Waypoint}         & 4.9  & 18.5 & 1.8  & 4.9  & 18.5 & 1.8  & 2.1  & 9.1  & 0.8  & 2.1  & 9.1  & 0.8 \\
Direction Follower~\cite{Waypoint}   & 54.7 & 72.0 & 41.1 & 11.1 & 17.2 & 8.4  & 32.3 & 41.2 & 23.8 & 13.9 & 18.0 & 10.7 \\
Frontier W~\cite{Waypoint}   & 44.0 & 63.9 & 35.2 & 6.5  & 14.8 & 5.1  & 30.6 & 42.8 & 22.2 & 10.9 & 16.4 & 8.1 \\
Supervised W~\cite{Waypoint} & 59.1 & 88.1 & 48.5 & 14.1 & 43.1 & 10.1 & 21.0 & 36.2 & 16.2 & 4.1  & 8.8  & 2.9 \\
Gan et al.~\cite{gan2020look}           & 57.6 & 83.1 & 47.9 & 7.5  & 15.7 & 5.7  & 22.8 & 37.9 & 17.1 & 5.0  & 10.2 & 3.6 \\
SoundSpaces~\cite{soundspaces}          & 74.4 & 91.4 & 48.1 & 34.7 & 50.9 & 16.7 & 54.3 & 67.7 & 31.3 & 21.9 & 33.5 & 10.4 \\
CRFN~\cite{CRFN} & 76.7 & 93.1 & 47.3 & 41.6 & 55.7 & 22.5 & 57.3 & 70.3 & 33.2 & 27.7 & 40.1 & 13.5 \\
AV-WaN~\cite{Waypoint} & \textbf{86.6} & \textbf{98.7} & \textbf{70.7} & 34.7 & 52.8 & 27.1 & 72.3 & 93.6 & 54.8 & 40.9 & 56.7 & 30.6 \\
\textbf{Samba(Ours)}         & 78.3 & 93.4 & 63.3 & \textbf{42.4} & \textbf{72.8} & \textbf{30.8} & \textbf{73.3} & \textbf{95.0} & \textbf{56.3} & \textbf{47.1} & \textbf{68.0} & \textbf{36.2} \\
\bottomrule
\end{tabular*}%
\end{table*}

We reformulate the state modeling operator of the backbone network into the Mamba State Encoder (M-SE) based on selective state-space models, marking a complete shift from the traditional gated recurrent logic to a modern sequential modeling paradigm. As illustrated in Fig. 3, the M-SE takes multimodal fused features $x_t \in \mathbb{R}^{d_{model}}$ from the perception layer and achieves real-time aggregation of navigation states in a fully data-driven manner. Before feeding into the core state-space operator, the input signal is first processed by a combination of linear projection and local 1D convolution, which effectively preserves local spatial correlations and feature smoothness during temporal sequence progression. Compared with the relatively fixed gating update mechanism of GRUs, the key advantage of M-SE lies in its selective mechanism: the parameters $\Delta$, $\mathbf{B}$, and $\mathbf{C}$ are no longer static weights, but functions dynamically generated conditioned on the current input $x_t$.

In terms of mathematical modeling, the underlying logic of M-SE originates from the discretization of continuous state-space equations. For the continuous hidden state $h(t)$, its evolution is governed by the linear differential equation $\dot{h}(t) = \mathbf{A}h(t) + \mathbf{B}x(t)$. To handle discrete navigation steps computationally, we apply the zero-order hold rule for discretization, thereby deriving the explicit expressions of the discrete transition matrix $\mathbf{\bar{A}}$ and input matrix $\mathbf{\bar{B}}$:

\begin{align}
\bar{\mathbf{A}} &= \exp(\Delta \mathbf{A}), \\
\bar{\mathbf{B}} &= (\Delta \mathbf{A})^{-1} \bigl( \exp(\Delta \mathbf{A}) - \mathbf{I} \bigr) \cdot \Delta \mathbf{B}.
\end{align}

In our practical lightweight PyTorch implementation, the above discretization procedure is further simplified to $\mathbf{\bar{A}} = \exp(\Delta \mathbf{A})$ and $\mathbf{\bar{B}} = \Delta \mathbf{B}$. Based on this, the recurrent update equation for the hidden state $\mathbf{h}_t \in \mathbb{R}^{d_{inner} \times d_{state}}$ can be formulated as:

\begin{equation}
\mathbf{h}_t = \bar{\mathbf{A}} \, \mathbf{h}_{t-1} + \bar{\mathbf{B}} \, x_t,
\end{equation}
this design endows the agent with stronger perceptual modeling capabilities from a mathematical essence. When the agent faces the challenge of information representation dilution during exploration, M-SE can dynamically control the decay rate of historical memory by adjusting the step size $\Delta$, thereby focusing on acoustic landmarks that are critical to target orientation. The computed hidden state is feature-mapped via the output parameter $\mathbf{C}$, then fused with the residual path through a gating mechanism. The final global hidden state $h_t$ serves as the core input to the decision-making system.

To balance the large-scale parallelism of policy training and the real‑time response requirements of embodied navigation, M‑SE supports dual‑mode switching between parallel and recurrent execution in its architecture. During training, the system employs a parallel scan algorithm for efficient gradient computation over long observation sequences, while during deployment, it seamlessly switches to single‑step recurrent mode to guarantee extremely low latency in decision‑making.

In the policy learning process, the system dynamically maintains the internal state matrices and local convolution buffers of the SSM through a specially designed state unpacking and packing mechanism. Particularly critical for adapting to the sampling nature of reinforcement learning, we introduce masking logic that forcibly resets Mamba’s internal hidden state at the end of each episode. This effectively prevents state contamination from prior experience on new tasks and ensures the stability of PPO optimization. Through such a comprehensive upgrade from operators to the underlying state machine, M‑SE provides Samba with far greater long‑range modeling depth than traditional recurrent architectures, significantly enhancing the agent’s perceptual robustness in complex and unknown environments.

\section{Experiments}

The experimental evaluation of this study mainly relies on the SoundSpaces acoustic simulation platform constructed on the Habitat simulator, and incorporates two widely adopted public datasets in the field of embodied intelligence namely Replica~\cite{replica} and Matterport3D(MP3D)~\cite{matterport3d}.
The MP3D dataset covers 85 real-world indoor environments dominated by residential spaces, providing the agent with complete 3D mesh models and corresponding panoramic image scanning data.
The Replica~\cite{replica} dataset contains 18 synthetic indoor scenes including apartments and hotels, which outperforms MP3D~\cite{matterport3d} in both the precision of geometric details and the realism of lighting effects.

To reconstruct complex acoustic propagation characteristics in virtual scenes, the SoundSpaces platform synthesizes physically realistic sound sources in real time by convolving selected audio signals with binaural room impulse responses at corresponding spatial positions.
In the specific experimental setup, we adopt 102 distinct natural sounds as source stimuli to ensure that the model has sufficient generalization ability when facing diverse acoustic inputs.
Through such a high-fidelity multimodal simulation environment, we are able to conduct a comprehensive and rigorous validation of the navigation performance of the Samba architecture under complex physical constraints.

In our experiments, two different sound source conditions are included.
(1) Heard: The target sound source is a telephone ringtone, which is used in the training, validation, and test sets.
(2) Unheard: The 102 sound sources are divided into three non-overlapping groups. Specifically, 78 sound sources are used for training scenes, 11 for validation scenes, and the remaining 18 for test scenes~\cite{CRFN}. All test scenes are unseen during training~\cite{CRFN}.

\subsection{Evaluation Metrics}
Following established protocols, the navigation performance is quantified via Success Rate ($SR$), Success weighted by Path Length ($SPL$), and Success Navigation Accuracy ($SNA$). $SR$ computes the fraction of runs where the agent successfully stops at the target. Trajectory optimality is captured by $SPL$ by balancing the shortest path $l_i$ against the execution path $p_i$, whereas strategic precision is monitored by $SNA$, which uses the action count $a_i$ to suppress redundant maneuvers like in-place rotations. We formulate these definitions as follows:
\begin{equation}\text{SR} = \frac{1}{N} \sum_{i=1}^{N} S_i\label{eq:sr},\end{equation}
\begin{equation}\text{SPL} = \frac{1}{N} \sum_{i=1}^{N} S_i \cdot \frac{l_i}{\max(p_i, l_i)}\label{eq:spl}~\cite{Waypoint},\end{equation}
\begin{equation}\text{SNA} = \frac{1}{N} \sum_{i=1}^{N} S_i \cdot \frac{l_i}{a_i}\label{eq:sna}~\cite{SPL},\end{equation}In these expressions, $N$ denotes the total number of test episodes while $S_i \in \{0, 1\}$ serves as the binary success indicator for the $i$-th episode. The variables $l_i$ and $p_i$ correspond to the shortest and actual path lengths respectively, and $a_i$ records the total count of actions executed throughout the navigation process.

\subsection{Quantitative Experimental Results}

As summarized in Table~\ref{table1}, Samba consistently outperforms all existing baselines across all test environments. On the Replica dataset~\cite{replica}, for instance, it achieves a 72.8\% success rate under the unheard-sound, unseen-scene setting, representing a 20 percentage point absolute improvement over the prior state-of-the-art method AV-WaN~\cite{Waypoint}. We also observe notable improvements in trajectory and decision efficiency: SPL climbs from 34.7\% to 42.4\%, and SNA grows from 27.1\% to 30.8\%. These simultaneous gains suggest that our selective scanning mechanism supports robust state tracking and extracts precise directional features from dynamic observation streams, which effectively reduces aimless wandering during navigation.

The framework also adapts well to the more acoustically complex MP3D dataset~\cite{matterport3d}. When tested with unheard sound sources, Samba achieves a 68.0\% SR, 47.1\% SPL, and 36.2\% SNA. Unlike conventional recurrent architectures that suffer from state dilution over long sequences, Samba maintains a better balance between audio and visual inputs. This result confirms that our redesigned backbone brings critical structural advantages for large-scale embodied navigation tasks.

\subsection{Efficiency Analysis}
\begin{table}[htbp]
\centering
\caption{Quantitative comparison and analysis between the Samba architecture and the AV-WaN baseline model in terms of audio encoding modules, state encoding modules, and total parameter sizes across different datasets. Lower values indicate improved model efficiency.}
\label{model_size}
\setlength{\tabcolsep}{4pt} 
\begin{tabular}{l|c|c|cc}
\toprule

\multirow{2}{*}{Method} & \multirow{2}{*}{Audio Enc (M)$\downarrow$} & \multirow{2}{*}{State Enc (M)$\downarrow$} & \multicolumn{2}{c}{Total (M)$\downarrow$} \\
\cmidrule(lr){4-5}

& & & Replica & MP3D \\

\midrule 
AV-WaN~\cite{Waypoint}       & 0.7 & 3.9 & 11.2 & 5.6 \\
Samba (Ours) & 0.4 & 2.7 & 10.5 & 4.6 \\

\bottomrule
\end{tabular}
\end{table}

Table~\ref{model_size} confirms that Samba achieves strong structural efficiency through its core operator design. The audio encoder contains only $0.4\text{M}$ parameters, a $43\%$ reduction from the $0.7\text{M}$ baseline, which demonstrates that a compact selective scanning mechanism captures global acoustic features more effectively than standard convolutional layers. We similarly trim state encoding parameters from $3.9\text{M}$ to $2.7\text{M}$ to cut architectural redundancy and maintain stable training when data is limited. Total parameter counts differ across datasets due to varying sensor resolutions, reaching $10.5\text{M}$ on Replica and $4.6\text{M}$ on MP3D, yet Samba consistently outperforms AV-WaN in both settings. Overall, our redesigned backbone preserves critical environmental cues without redundant network stacking, and adapts the framework well for edge robotics deployment.

\subsection{Ablation Studies}

Ablation results in Table~\ref{xr1} and Table~\ref{xr2} evaluate the incremental gains of our Mamba-based components, where “w/o AME” and “w/o M-SE” represent reverting the respective modules back to the baseline CNN and GRU architectures. Replacing AME triggers a substantial performance collapse, with $SR$ for unheard sounds dropping from $72.8\%$ to $59.0\%$ on Replica and from $68.0\%$ to $62.6\%$ on MP3D. Conversely, the configuration without M-SE results in relatively moderate fluctuations, with Replica $SR$ maintaining $71.0\%$. These findings confirm that perceptual precision is the dominant driver of success, validating our perception-centric architectural design.

\begin{table}[htbp]
\centering
\caption{Ablation study on the Fusion Controller of CRFN on the Replica dataset.}
\label{xr1}
\setlength{\tabcolsep}{6pt} 
\renewcommand{\arraystretch}{1.15}

\begin{tabular}{l|ccc|ccc}
\toprule
\multirow{2}{*}{\textbf{Method}}  
  & \multicolumn{3}{c|}{\textbf{Heard}}      
  & \multicolumn{3}{c}{\textbf{Unheard}} \\
\cmidrule(lr){2-4} \cmidrule(lr){5-7}
& \textbf{SPL} $\uparrow$ & \textbf{SR} $\uparrow$ & \textbf{SNA} $\uparrow$
& \textbf{SPL} $\uparrow$ & \textbf{SR} $\uparrow$ & \textbf{SNA} $\uparrow$ \\
\midrule
w/o AME           & 77.9 & 92.2 & 63.3 & 40.0 & 59.0 & 31.4 \\
w/o M-SE           & 82.5 & 97.3 & 66.5 & 44.1 & 71.0& 33.7 \\
\textbf{Samba(Full)} & 78.3 & 93.4 & 63.3 & 42.4 & 72.8 & 30.8 \\
\bottomrule
\end{tabular}
\end{table}

\begin{table}[htbp]
\centering
\caption{Ablation study on the Fusion Controller of CRFN on the Matterport3D dataset.}
\label{xr2}
\setlength{\tabcolsep}{6pt} 
\renewcommand{\arraystretch}{1.15}

\begin{tabular}{l|ccc|ccc}
\toprule
\multirow{2}{*}{\textbf{Method}}  
  & \multicolumn{3}{c|}{\textbf{Heard}}      
  & \multicolumn{3}{c}{\textbf{Unheard}} \\
\cmidrule(lr){2-4} \cmidrule(lr){5-7}
& \textbf{SPL} $\uparrow$ & \textbf{SR} $\uparrow$ & \textbf{SNA} $\uparrow$
& \textbf{SPL} $\uparrow$ & \textbf{SR} $\uparrow$ & \textbf{SNA} $\uparrow$ \\
\midrule
w/o AME           & 72.1 & 93.6 & 54.5 & 41.8 & 62.6 & 31.6 \\
w/o M-SE           & 71.8 & 93.5 & 55.6 & 42.8 & 60.2 & 32.1 \\
\textbf{Samba(Full)} & 73.3 & 95.0 & 56.3 & 47.1 & 68.0 & 36.2 \\
\bottomrule
\end{tabular}
\end{table}

\section{CONCLUSIONS}

We propose Samba, a hybrid architecture built on state-space models, to overcome the long-standing efficiency and modeling bottlenecks in audio-visual navigation. Samba integrates an adaptive selective Mamba State Encoder with a bidirectional Audio Mamba Encoder; this combination successfully captures global time-frequency correlations where convolutions fail and prevents the representation dilution typical of recurrent units. Our empirical validation shows that updating the backbone not only yields stronger embodied representations but also slashes parameter size and computation. This shift provides an efficient, high-performance reference design for future embodied AI deployment.

\addtolength{\textheight}{-12cm}   




\bibliographystyle{IEEEtran} 
\bibliography{references}

\end{document}